\documentclass[letterpaper, 10 pt, conference]{ieeeconf}  % Comment this line out if you need a4paper

\IEEEoverridecommandlockouts                              % This command is only needed if 
\usepackage{cite}
\usepackage{amsmath,amssymb,amsfonts}
\usepackage{multirow} 
\usepackage{algorithmic}
\usepackage{graphicx}
\usepackage{textcomp}
\usepackage{tabularx}
\usepackage{booktabs}
\usepackage{setspace}
\usepackage{arydshln} % Add this to your preamble
\usepackage{array}
\usepackage{boldline}
\usepackage{float}
\usepackage{subcaption}
\usepackage{tabularx}
\usepackage{xcolor}
\def\BibTeX{{\rm B\kern-.05em{\sc i\kern-.025em b}\kern-.08em
    T\kern-.1667em\lower.7ex\hbox{E}\kern-.125emX}}

\title{\LARGE \bf
AGRNav: Efficient and Energy-Saving Autonomous Navigation for Air-Ground Robots in Occlusion-Prone Environments
}

\author{Junming Wang$^{1}$, Zekai Sun$^{1}$, Xiuxian Guan$^{1}$, Tianxiang Shen$^{1}$, Zongyuan Zhang$^{1}$, \\Tianyang Duan$^{1}$, Dong Huang$^{1}$, Shixiong Zhao$^{3}$, Heming Cui$^{1,2,*}$% <-this % stops a space
\thanks{*denotes corresponding author.}% <-this % stops a space
\thanks{$^{1}$J Wang, Z Sun, X Guan,  T Shen, Z Zhang, T Duan, D Huang, H Cui is with the University of Hong Kong. $^{2}$H Cui is with the Shanghai AI Laboratory. $^{3}$S Zhao is with the Huawei Technologies, Co. Ltd.}%
}

\begin{document}
\maketitle
\thispagestyle{empty}
\pagestyle{empty}
%%%%%%%%%%%%%%%%%%%%%%%%%%%%%%%%%%%%%%%%%%%%%%%%%%%%%%%%%%%%%%%%%%%%%%%%%%%%%%%%
\begin{abstract}

The exceptional mobility and long endurance of air-ground robots are raising interest in their usage to navigate complex environments (e.g., forests and large buildings). However, such environments often contain occluded and unknown regions, and without accurate prediction of unobserved obstacles, the movement of the air-ground robot often suffers a suboptimal trajectory under existing mapping-based and learning-based navigation methods. In this work, we present AGRNav, a novel framework designed to search for safe and energy-saving air-ground hybrid paths. AGRNav contains a lightweight semantic scene completion network (SCONet) with self-attention to enable accurate obstacle predictions by capturing contextual information and occlusion area features. The framework subsequently employs a query-based method for low-latency updates of prediction results to the grid map. Finally, based on the updated map, the hierarchical path planner efficiently searches for energy-saving paths for navigation. We validate AGRNav's performance through benchmarks in both simulated and real-world environments, demonstrating its superiority over classical and state-of-the-art methods. The open-source code is available at https://github.com/jmwang0117/AGRNav.

\end{abstract}

%%%%%%%%%%%%%%%%%%%%%%%%%%%%%%%%%%%%%%%%%%%%%%%%%%%%%%%%%%%%%%%%%%%%%%%%%%%%%%%%
\section{INTRODUCTION}

Air-ground robots (AGR), which are known for their outstanding mobility and long endurance, have been gaining significant interest lately and show great potential for applications in search and rescue tasks \cite{kalantari2013design, pan2023skywalker,qin2020hybrid}. Existing works \cite{zhang2022autonomous, fan2019autonomous,suh2020energy} have demonstrated success in fast air-ground hybrid path planning, particularly in simple and unobstructed scenarios. However, AGR navigating complex environments (e.g., forests or buildings) with occluded and unknown areas faces a dilemma since obstacles in these areas significantly affect the results of path planning, i.e., high collision probability and suboptimal energy consumption (in Fig.1a).

To enable efficient and energy-saving navigation for air-ground robots in occluded environments, existing \textbf{\textit{mapping-based}} methods \cite{fan2019autonomous, zhang2022autonomous} use sensors (e.g., cameras or LiDAR) to construct a local occupancy grid map and an Euclidean Signed Distance Field (ESDF) map \cite{zhou2019robust} for fast path planning. However, since the sensors' limitation is perceiving only visible obstacles (in Fig.1a), the constructed maps exclude obstructions in occluded areas, which increases the risk of collisions and leads to higher energy consumption from unnecessary aerial paths.

In contrast, existing \textbf{\textit{learning-based}} methods employing semantic scene completion networks \cite{cao2022monoscene, li2023voxformer,song2017semantic} to predict obstacle distribution in occluded areas and then enable the path planner to reduce unnecessary paths to achieve energy savings. Some networks use 3D convolutions \cite{tran2015learning} for enhanced prediction accuracy; however, their memory-intensive nature and high inference latency make them unsuitable for real-time robotic applications. While some work \cite{wang2021learning, roldao2020lmscnet} focuses on developing lightweight networks and achieving success in real-time inference, the network's limited ability to capture features and contextual information makes its prediction accuracy drop significantly. Additionally, addressing the update delay issue is also important, as delays may cause the path planner to ignore predicted obstacle distribution, thereby leading to similar problems as in mapping-based methods. 

\begin{figure}[t]
  \centering
     \includegraphics[width=\linewidth ]{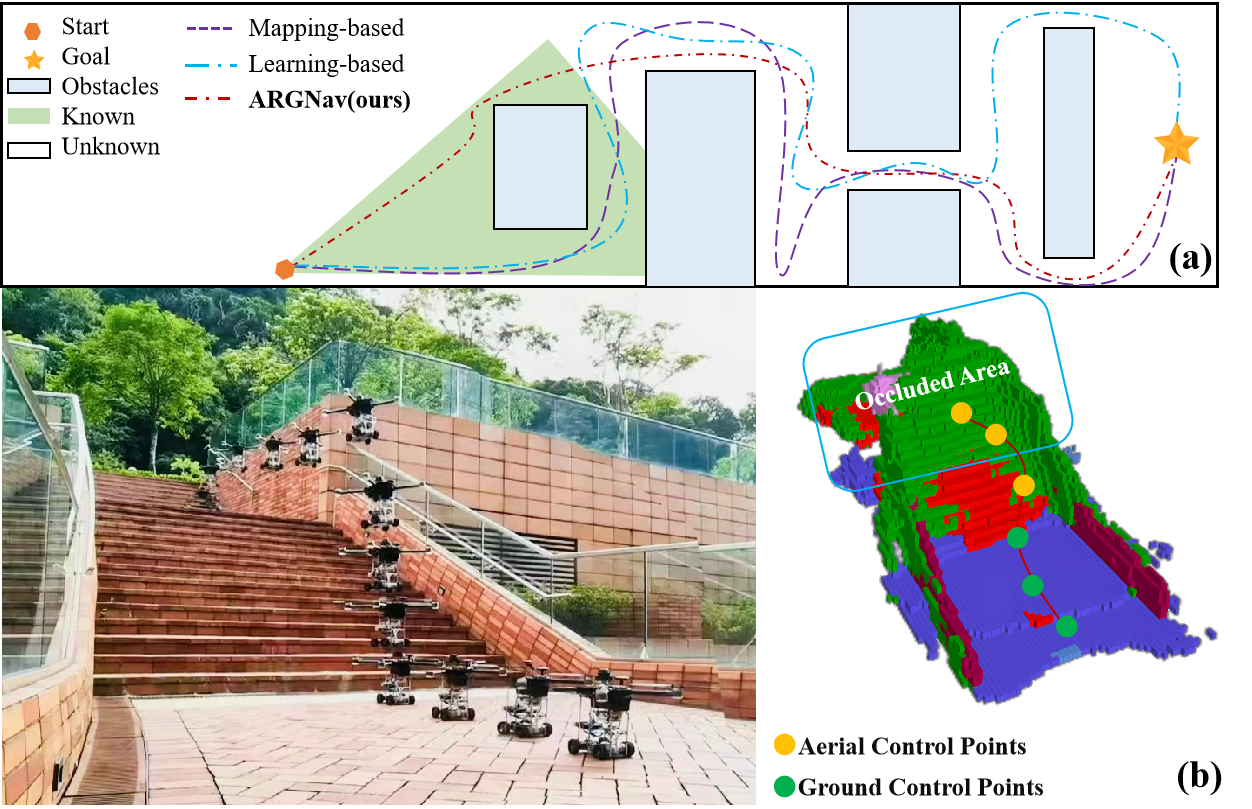}
     \caption{\small{\textbf{(a)} Previous navigation systems had problems predicting occlusions, resulting in higher collision probabilities and suboptimal pathways that consumed more energy.  \textbf{(b)} By predicting occlusions in advance, AGRNav can minimize and avoid collisions, resulting in efficient and energy-saving paths.}}
\end{figure}

\begin{figure}[t]
  \centering
     \includegraphics[width=0.95\linewidth]{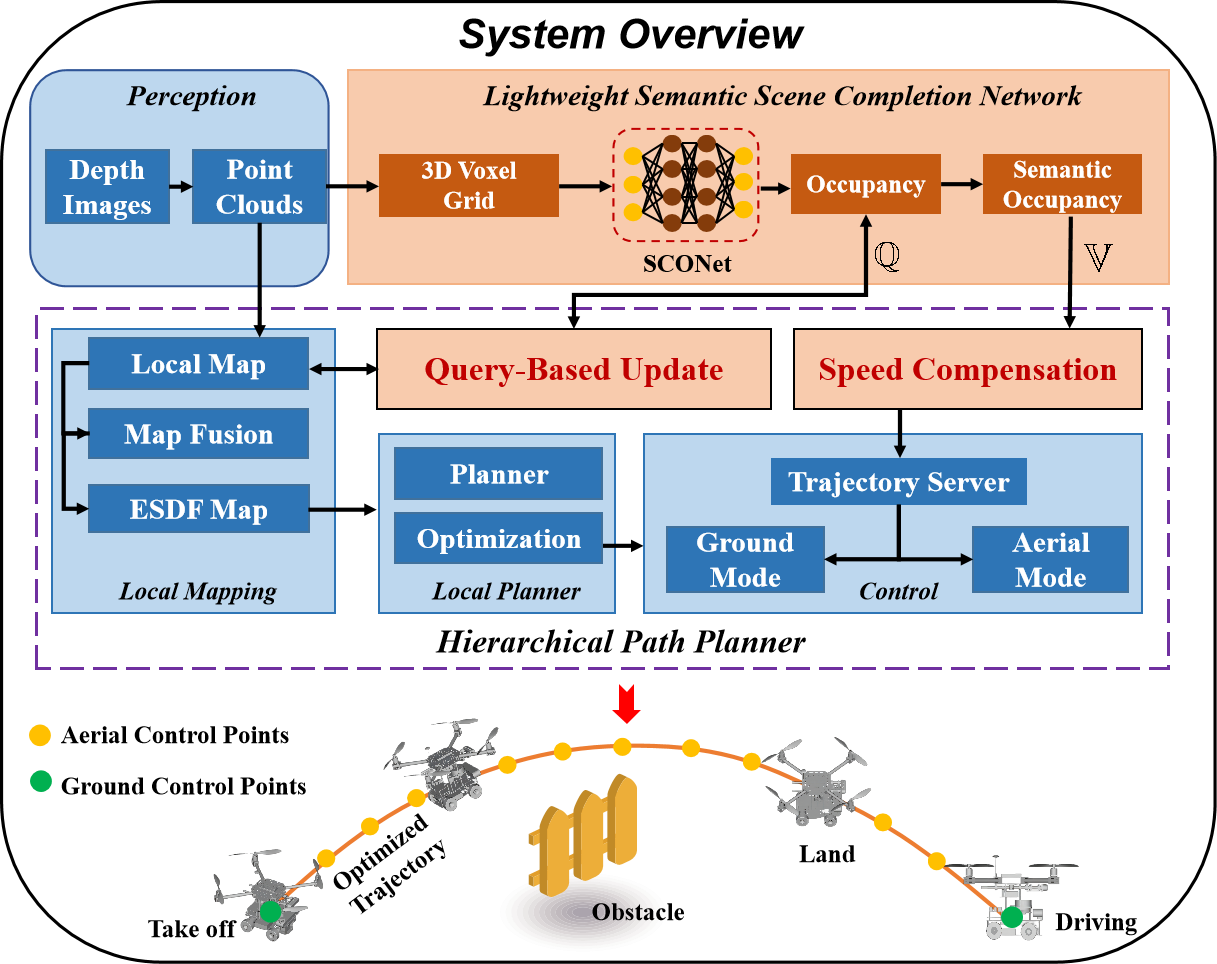}
   \caption{\raggedright\small The overview of our proposed Framework: AGRNav. $\mathbb{Q}$ denotes that the free voxels in the grid map query and update their occupancy status from the predicted occupancy map. $\mathbb{V}$ denotes that predicted semantics is turned into speed compensation.}
   \label{fig:onecol}
\end{figure}

To tackle the high inference latency of memory-intensive networks and the low prediction accuracy of lightweight networks due to their inability to capture useful features, our key observation involves integrating lightweight convolutions and self-attention mechanisms into the network. The former allows the network to perform real-time inference tasks on robotic devices, while the latter enhances the network's ability to learn long-distance dependencies and capture contextual information, which is beneficial for improving the accuracy of prediction. Moreover, regarding update delay issues in existing methods that depend on map merging and result in repeated updates of occupied voxels, one potential method is only querying and updating the occupancy status of free voxels after scanning to ensure low latency.

Based on the above observations, we present \textbf{\textit{AGRNav}}, a novel efficient and energy-saving path-planning framework.  The framework consists of two key components, the first one is a lightweight semantic scene completion network (SCONet), which is deployed on AGR and performs fast inference to accurately predict obstacle distribution and semantics. SCONet processes 3D voxel grids using depth-separable convolutions \cite{chollet2017xception} rather than 3D convolutions, which greatly decreases the number of calculations. Furthermore, to enable SCONet to capture rich and dense contextual information as well as features of occlusion areas, it integrates two self-attention mechanisms. This keeps the network lightweight while enhancing its feature extraction capabilities (Fig.1b).

The hierarchical path (i.e., aerial and ground path) planner (in Fig. 2) utilizes a query-based method for low-latency occupancy updates. With the accurate predictions of SCONet, the planner minimizes collisions and energy consumption while searching for paths on an updated map containing scanned and predicted obstacles. Furthermore, it offers speed compensation for the robot using the semantics predicted, allowing for acceleration in passable areas, e.g., roads.

Simulations and real-world experiments show that the AGRNav enable search for safe and energy-saving pathways in occlusion-prone environments. The following are the key contributions of this paper:
\begin{itemize}
\item \textbf{AGRNav is efficient.} AGRNav achieves a 98\% success rate in occlusion environments while also being low-latency in updating prediction results to the grid map.
\item \textbf{AGRNav is energy-saving.} By predicting obstacle distribution in advance, unnecessary aerial paths are substantially reduced, resulting in a 50\% decrease in energy consumption compared to the baseline.
\item \textbf{SCONet is lightweight and accurate.} SCONet enables real-time (i.e., 20 FPS) and accurate inference and achieves state-of-the-art performance (IoU = 56.12) on the SemanticKITTI benchmark.
\end{itemize}

\section{Related Work}

\subsection{Autonomous Navigation of Air-Ground Robots}

The escalating interest in the adaptability and versatility of AGR has led to a surge of research and innovations in the field. Although many researchers prioritize mechanical structure design \cite{zhang2022autonomous, qin2020hybrid,fan2019autonomous,pan2023skywalker} to minimize weight and volume, it is crucial to acknowledge that establishing an efficient and energy-saving navigation framework that empowers AGR to navigate in complex environments carries greater significance. Despite this, there remains room for improvement and further investigation in current air-ground robot navigation frameworks. For example, \cite{fan2019autonomous} presented air-ground path planning work, but due to the absence of trajectory refinement methods, the resulting trajectories lack smoothness and dynamic feasibility. \cite{zhang2022autonomous} proposed an energy-saving and fast autonomous navigation framework, but its ``aggressive" planning strategy increases the risk of collision when navigating complex and occluded areas.

\subsection{Navigation in Predicted Maps}

Autonomous navigation with low collision probability and energy savings by predicting obstacle distribution in occluded areas has shown promising results in recent studies. However, existing methods face limitations in complex environments and high-speed navigation scenarios. For instance, \cite{katyal2021high} introduces novel perception algorithms and a controller that incorporate predicted occupancy maps for high-speed navigation. Despite its potential, the method struggles to handle complex and obstacle-dense environments due to simplistic scene design and a lower map update frequency ($\approx$ 3 Hz). Similarly, \cite{elhafsi2020map} employs a conditional neural process-based network to predict map turns but relies on heuristic approaches for motion planning in unknown environments. This results in greedy and inefficient trajectories without considering the unobserved environment's structure. Lastly, \cite{wang2021learning} proposed OPNet, a method that predicts occupancy grids for path planning and performs well in simple environments. However, the method faces challenges in large-scale occluded scenes because its network does not have the ability to capture the characteristics and contextual information of occluded areas.

\subsection{Semantic Scene Completion and Occupancy Mapping}

Robot sensors with narrow fields of view, such as LiDAR and depth cameras, make it difficult to monitor occluded areas. The majority of the research on predicting occluded region occupancy using limited sensor data has focused on semantic scene completion approaches. Notable works include SSCNet \cite{song2017semantic} by \textit{Song et al.,} which uses depth images to predict occupancy and semantics for voxels. Monoscene \cite{cao2022monoscene} by \textit{Cao et al.,}  only requires monocular RGB images and leverages a novel 2D-3D feature projection bridge to predict occupancy and semantics for voxels. However, these memory-intensive methods are unsuitable for real-time inference on robots' devices since Monoscene \cite{cao2022monoscene} and VoxFormer's \cite{li2023voxformer} GPU memory exceeds 10 GB during inference. 

\begin{figure*}[t]
  \centering
  \includegraphics[width=0.9\linewidth]{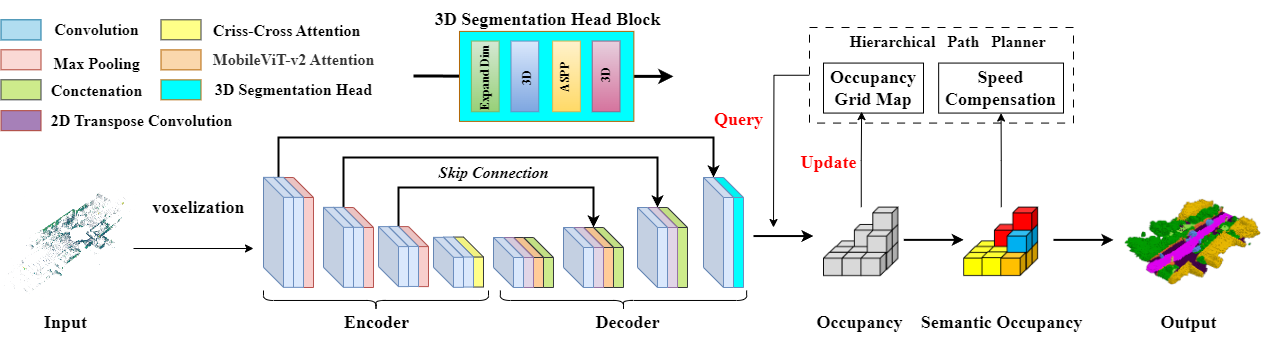}
  \caption{\small\textbf{SCONet: Lightweight Semantic Scene Completion Network.} Our network employs a self-attention-driven U-Net architecture, featuring depthwise separable convolutions and segmentation heads, to perform efficient 3D scene completion and semantic segmentation.}
   \label{fig:onecol}
\end{figure*}

\section{System Overview}

Fig.2 illustrates the proposed framework \textbf{\textit{AGRNav}}, featuring some key components: (1) the lightweight semantic scene completion network SCONet (Section IV); (2) the query-based low-latency occupancy update method (Section V-A); (3) the hierarchical path planner (Section V-B) search air-ground hybrid paths on the updated map which contains scanned obstacles and predicted obstacles. 

\section{Semantic scene completion network}

\subsection{SCONet Network Structure}

We proposed a lightweight semantic scene completion network (SCONet) to predict the distribution of obstacles in occluded regions, as shown in Fig. 3. Point clouds are transformed into a 3D sparse voxel grid, serving as input for our 4-level U-Net style network.  Each voxel is assigned a semantic label $\mathcal{L} = [i,l_{1},l_{2},...,l_{N}], i=0,1$, where $N$ is the number of semantic classes, $i = 0$ represents free voxels, and $i = 1$ represents occupancy voxels. This design allows for the effective prediction of obstacle distribution and their corresponding semantics using partial scans. The specific encoder and decoder structures are as follows:

\noindent\textbf{Encoder.} Instead of using memory-intensive 3D convolutions \cite{tran2015learning}, we use considerably lighter depthwise separable convolutions \cite{chollet2017xception} along the X and Y dimensions in the encoder, changing the height dimension (Z) into a feature dimension. This design learns features at a lower resolution while allowing direct processing of 3D voxel grids.

\noindent\textbf{Decoder.} By employing deconvolutions in the decoder, we up-sample the feature maps and subsequently concatenate output results to lower levels, which enhances the information flow while enabling our network to learn high-level features from coarser resolutions. Lastly, the semantics will be predicted through a 3D segmentation block that has a series of dense and dilated convolutions \cite{yu2015multi}.

\subsection{Two GPU Memory-Efficient Self-attention Mechanisms}

The above design makes SCONet suitable for deployment on robotic devices for real-time inference. However, the network lacks the ability to capture contextual information and features in occluded areas, which is essential for improving prediction accuracy in such areas. Therefore, we have integrated two self-attention mechanisms into our architecture: \textit{Criss-Cross Attention (CCA)} \cite{huang2019ccnet} and \textit{MobileViT-v2 Attention} \cite{mehta2022separable}. CCA, which is positioned after the encoder (in Fig. 3), enhances the network's ability to learn long-distance dependencies by collecting contextual information in horizontal and vertical directions. This enables the establishment of connections between distant features, which leads to more effective predictions of obstacles and semantics in occluded environments by comprehending the relationships among various elements (e.g., roads and walls).

Specifically, CCA takes the feature map $\mathcal{\mathbf{H} }\in \mathbb{R}^{C\times W\times H}$ output from the fourth convolutional layer of the encoder as input, and then through two convolutional layers with $1 \times 1$ filters to acquire feature maps $Q$, $K$ and attention maps $A$ via affinity operation. Affinity can be defined as follows:
\begin{equation}
  \small  d_{i,u}=\mathbf{Q_u} \mathbf{\Omega _i,u^T} 
\end{equation}
where  ${\mathbf{Q, K} } \in \mathbb{R}^{C^{'} \times W\times H} $, $\mathbf{A} \in \mathbb{R}^{(H+  W- 1)\times (W\times H)} $, $Q_u$ is a vector at each position $u$ in the spatial dimension of feature maps $Q$, $\Omega _u$ is a set combined all feature vectors from $\mathrm{K}$ which are in the same row or column with position $u$. The contextual information is collected by an aggregation operation defined as follows:
\begin{equation}
   \small \mathbf{H} _{u}^{'}=\sum_{i=0}^{H+W-1}  \mathbf{A} _{i,u}\mathbf{\Phi}  _{i,u}+\mathbf{H} _u
\end{equation}
where $\mathbf{H} _{u}^{'}$ is a feature vector at position $u$ and $i$ means channel. The contextual information is added to local feature $H$ to augment the voxel-wise representation. The CCA's ability to learn long-distance dependencies enables SCONet to effectively understand the context and relationships between various structural elements in the environment, resulting in accurate obstacle prediction. To further achieve finer-grained semantic scene completion, such as trees and cars, and better capture features of regions in occluded areas, which in turn contributes to the reduction of robot collision probability, we integrated MobileViT-v2 Attention \cite{mehta2022separable} into the first (resolution 1:8) and second (resolution 1:4) layers of SCONet's decoder. This integration allows the extraction of diverse resolution fine-grained features, which further enhances the completion of areas of occluded regions, i.e., improves obstacle prediction accuracy. With a latency of just 3.4 ms \cite{mehta2022separable}, MobileViT-v2 Attention allows SCONet to maintain stronger feature capture capabilities while remaining lightweight. Mathematically, MobileViT-v2 Attention \cite{mehta2022separable} can be defined as:
\begin{equation}
  \footnotesize {\scriptsize  \mathbf{y} =\left (\underbrace{ \sum \left ( \overbrace{\sigma (\mathbf{x} W_I)}^{c_\mathbf{s} \in \mathbb{R}^k } *\mathbf{x} W_K \right )}_{\substack{c_\mathbf{v} \in \mathbb{R}^d }} * \operatorname{ReLU}(\mathbf{x} W_V)  \right )W_O  } 
\end{equation}
where $\mathbf{x}$ as input and $*$ means broadcastable element-wise multiplication and $\sum$ means summation operations. $W_O\in \mathbb{R}^{d\times d}$ means linear layer with weights. A ReLU activation to produce an output $\mathbf{x} _V\in \mathbb{R}^{k\times d} $.

\section{Safe Air-Ground Hybrid Path Planner}

The hierarchical path planner, building on the aerial-ground integration proposed by Zhang et al. \cite{zhang2022autonomous}, adeptly merges a query-based occupancy update mechanism, kinodynamic trajectory searching methodologies, and a gradient-based spline optimizer. Our hierarchical planner facilitates the creation of energy-efficient hybrid trajectories and enhances overall planning efficiency.

\subsection{Query-Based Low-Latency Occupancy Update}

The SCONet network generates a predicted occupancy grid map with occupied and free voxels. Typically, this map is merged with scan-based occupancy grid maps to construct the ESDF map for planning. The time complexity of this merge operation is  $O(N)$, where N is the number of voxels since it needs to traverse and combine information from both grid maps. To achieve efficient navigation and obstacle avoidance, we proposed a query-based update method with low latency. Specifically, $f(x, S_{\text{pred}})$ represents the query operation, which checks whether the voxel $x$ exists within the predicted occupied voxel set $S_{\text{pred}}$. If $x$ is predicted to be occupied (i.e., $x \in S_{\text{pred}}$), then $f(x, S_{\text{pred}}) = \text{occupied}$; otherwise, the status of $x$ remains free. By focusing on $M$ relevant free voxels, where $M \leq N$, this method reduce the time complexity to $O(M)$.

\begin{equation}
S_{\text{updated}}(x) = \begin{cases} 
\text{occupied}, & \text{if } f(x, S_{\text{pred}}) = \text{occupied} \\
\text{free}, & \text{otherwise}
\end{cases}
\end{equation}

\subsection{Efficient and Energy-saving Hierarchical Path Planner}
Different from the rough path search method of \textit{Fan et al.} \cite{fan2019autonomous}, we also further optimize the trajectories (contains ground and aerial trajectories), that is, set the trajectories as a $p_b$ degree uniform B-spine with control points $\mathcal{\mathbf{P} }  =\left \{ \mathbf{P}_0, \mathbf{P}_1, \mathbf{P}_2,..., \mathbf{P}_N\right \}$. In particular, the optimization and generation of trajectories are mainly divided into ground and aerial trajectories. When optimizing ground trajectories, we assume that the AGR moves on flat ground, so we only need to consider the two-dimensional motion control point, denoted as:
\begin{equation}
  \small \mathcal{\mathbf{P} }_g  = \left \{ \mathbf{P}_{t0} ,\mathbf{P}_{t1} , \mathbf{P}_{t2} ,\mathbf{P}_{t3} ,..., \mathbf{P}_{tM-1} , \mathbf{P}_{tM} \right \}
\end{equation}
where $\mathbf{P} _{ti}=(x_{ti},y_{ti}),i\in [0, M]$. Meanwhile, the aerial trajectory control points are denoted as: $\mathcal{\mathbf{P} }_a$. 
We also use the following cost terms designed by \textit{Zhou  et al.} \cite{zhou2019robust} to refine the trajectory:
\begin{equation}
  \small  f_1=\lambda _sf_s+\lambda _cf_c+\lambda _f(f_v+f_a)
\end{equation}
where $\lambda _s, \lambda _c, \lambda _f$ are weights for each cost terms. $f_s$, $f_c$, $f_v$ and  $f_a$ are smoothness, collision cost,  soft limits on velocity and acceleration. We set the AGR to move in the ground mode, its speed is parallel to the yaw angle. In addition, considering that our AGR adopts the Akaman structure if the trajectory is too curved, there will be a huge error, so we enforce a cost on $\mathcal{\mathbf{P} }_g$ to limit the curvature of the terrestrial trajectory, the curvature at $\mathcal{\mathbf{P} } _{ti}$ is defined as:
\begin{equation}
 \small  \mathcal{C} _i=\frac{\bigtriangleup \beta _i}{\mathcal{P} _{ti}} 
\end{equation}
where 
$\footnotesize   \bigtriangleup \beta _i=|tan^{-1}\frac{\bigtriangleup y_{ti+1}}{\bigtriangleup x_{ti+1}}-tan^{-1}\frac{\bigtriangleup y_{ti}}{\bigtriangleup x_{ti}} | $. Therefore, this cost can be formulated as:
\begin{equation}
 \small  f_n=\sum_{i=1}^{M-1}F_n({\mathbf{P} }_{ti}) 
\end{equation}
Lastely,  the overall objective function is formulated as follows:
\begin{equation}
 \small    f_{total}=\lambda _sf_s+\lambda _cf_c+\lambda _f(f_v+f_a)+\lambda _nf_n
\end{equation}
and we use a non-linear optimization solver $NLopt^{2}$ to solve this optimization problem. After path planning is completed, a setpoint on the generated trajectory is selected according to the current timestamp and then sent to the controller. The settings and selections of the aerial and ground setpoint are the same as in \cite{zhang2022autonomous}.

\section{Experiments}

We evaluate AGRNav's improvement by comparing it against two mapping-based approaches and one learning-based method in two simulated environments. Moreover, we test AGRNav in three complex real-world scenarios employing a custom robot, showcasing its energy-saving advantages in practical navigation. By documenting the average energy consumption per second for AGR amidst driving and flying, we also establish a foundation for energy usage evaluation in simulated tests. Ultimately, we analyze SCONet's accuracy and real-time performance on the SemanticKITTI dataset.

\subsection{Simulated Air-Ground Robot Navigation} 
The simulated experimental setup includes a $20m \times  20m \times 5m$ square room and a $3m \times 30m \times 5m$ corridor, which are filled with random obstacles, leading to numerous occlusion spaces and unknown regions throughout the scene. The air-ground robot must navigate from the starting point to the destination, and the maximum speed does not exceed 2.5 m/s.

\begin{figure}[t]
  \centering
     \includegraphics[width=0.9\linewidth]{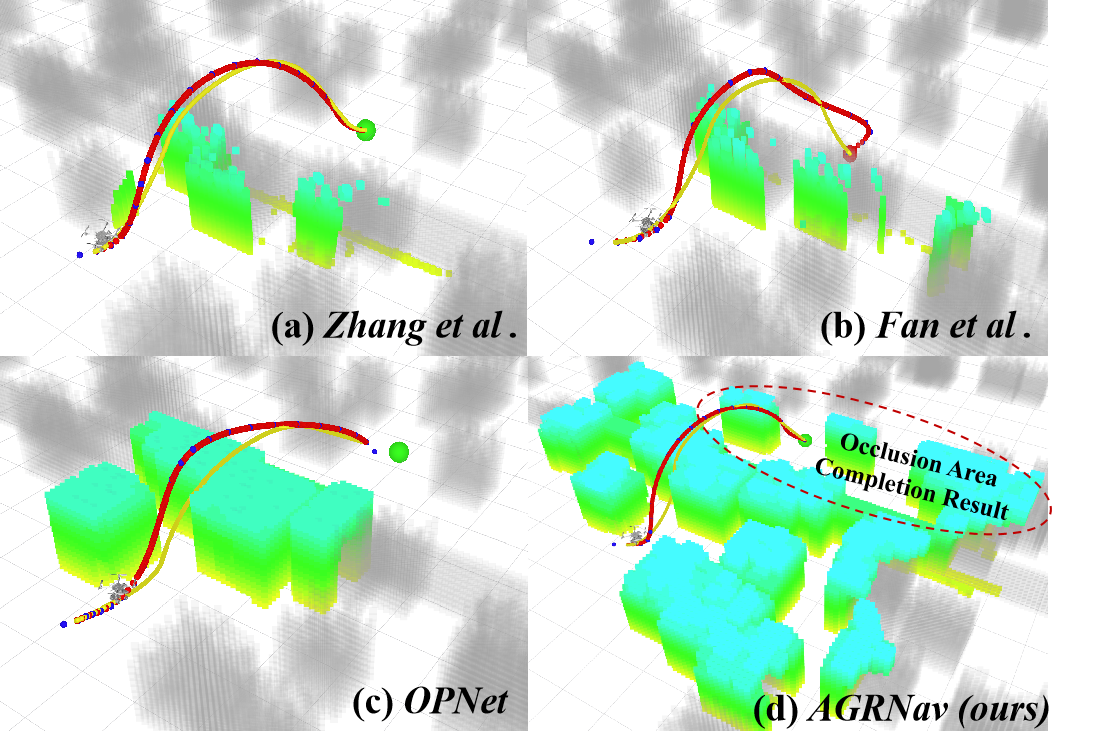}
   \caption{\small Four methods were used to plan paths in a simulated square room. AGRNav demonstrates the ability to predict the distribution of obstacles in occluded areas. }
   \label{fig:onecol}
\end{figure}

\begin{table}[ht]
    \centering
    \footnotesize
    \renewcommand{\arraystretch}{1.3} % Set row spacing to 1.2
    \caption{\small Quantitative results in two simulation scenarios.}
    \setlength{\tabcolsep}{2.2pt} % Reduce column spacing
    \begin{tabular}{cccccc}
    \toprule
    \textbf{Env.} & \textbf{Method} & \textbf{Succ. (\%)} & \textbf{Time (s)} & \textbf{Leng. (m)} & \textbf{Power (W)} \\
    \midrule
                             & Fan's \textcolor{green}{\cite{fan2019autonomous}} & 85.0 & 13.13 & 33.79 & 919.07 \\
    \textbf{\textit{Square}} & Zhang's \textcolor{green}{\cite{zhang2022autonomous}} & 95.0 & 12.05 & 23.09 & 793.30 \\
    \textbf{\textit{Room}}   & OPNet \textcolor{green}{\cite{wang2021learning}} & 91.0 & 12.90  & 32.12  & 888.04\\
                             & \textbf{AGRNav(Ours)} & \textbf{98.0} & \textbf{11.02}  & \textbf{21.82} & \textbf{434.55}\\
    \cdashline{1-6}
                                & Fan's \textcolor{green}{\cite{fan2019autonomous}} & 88.0 & 21.24 & 33.10  & 565.24\\
    \textbf{\textit{Corridor}}  & Zhang's \textcolor{green}{\cite{zhang2022autonomous}} & 97.0 &\textbf{16.97}  & 30.69 &519.20 \\
                                & OPNet \textcolor{green}{\cite{wang2021learning}} & 90.0 & 18.45  & 32.85& 534.11  \\
                                & \textbf{AGRNav(Ours)} & \textbf{98.0} & 17.50 & \textbf{29.82} & \textbf{445.61} \\
    \bottomrule
    \end{tabular}
    \label{tab:empty_table}
\end{table}

\noindent\textbf{Quantitative Results.} We conducted a comparative analysis of our AGRNav navigation framework against two mapping-based and one learning-based navigation method in a square room and corridor scenario. 100 trials with varying obstacle placements, we recorded the average travel time, length and success rate (i.e., no collisions) for all 4 methods. In particular, the energy consumption of the four methods is calculated using the energy consumed per second by our customized robot when flying and driving in the real environment (Table 2). Table 1 shows that our AGRNav outperforms the other three approaches, achieving the highest success rate (i.e., 98\%), since our network (SCONet) predicts a broader range of occlusion areas (in Fig.4d), and generates the path with the lowest collision rate. 

Furthermore, our framework substantially reduces redundant paths and cuts energy consumption by half (i.e., average consumption per second is 434.55 W) in a square room. This efficiency stems from SCONet's accurate predictions, which minimize high-energy-consuming aerial paths in favour of low-energy ground paths. In the corridor scene, while the average travel time of \cite{zhang2022autonomous} is shorter (i.e., 16.97 s), its average energy consumption is higher due to the inability to predict occlusion areas and a greater reliance on aerial paths.

\begin{figure}[t]
  \centering
     \includegraphics[width=0.9\linewidth]{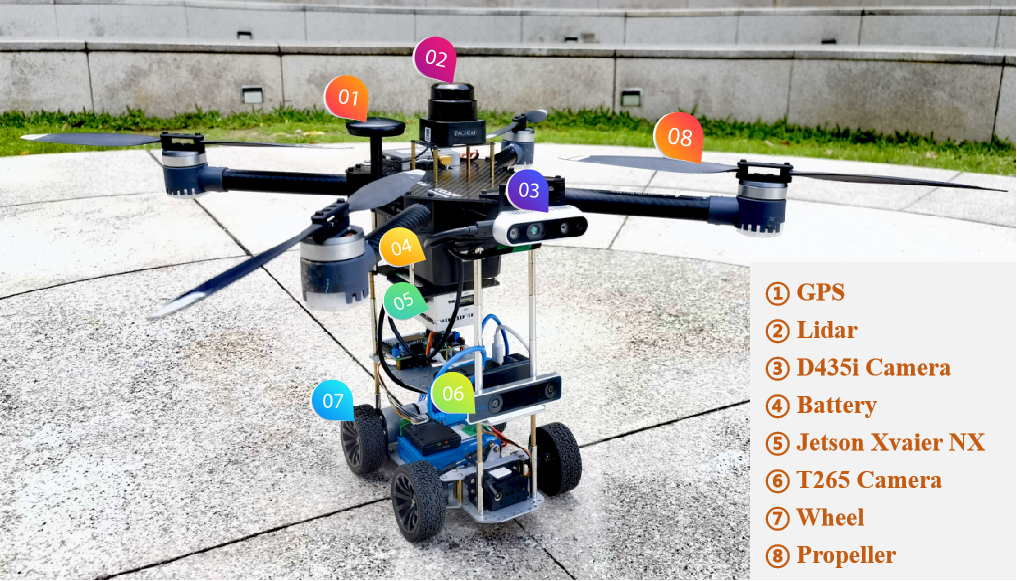}
   \caption{\small The detailed composition of our customized air-ground robot (AGR).}
   \label{fig:onecol}
\end{figure}

\subsection{Real-world Air-Ground Robot Navigation}
Our custom AGR platform (Fig. 5), is composed of a quadrotor with a 600mm diagonal wheelbase. This platform employs the Prometheus \cite{Prometheus} software system and is equipped with a RealSense D435i depth camera and a T265 camera. It also features a Jetson Xavier NX onboard computer for the deployed AGRNav framework. Mobility is sustained by a 10,000 mAh energy source, which enables up to 26 minutes of hovering. Table 2 shows the energy consumption per second in different modes. 

\begin{figure}[t]
  \centering
     \includegraphics[width=0.9\linewidth]{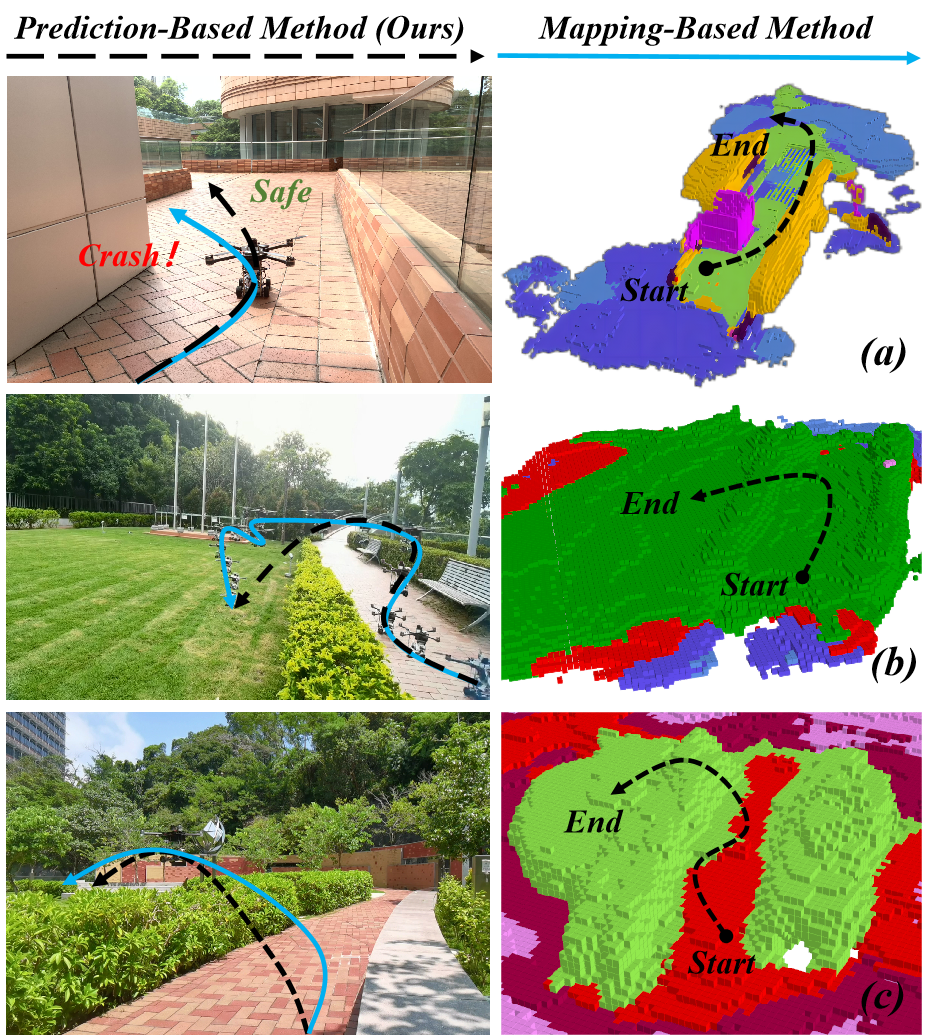}
   \caption{\small Navigation experiments of AGR in 3 complex real environments.}
   \label{fig:onecol}
\end{figure}

\begin{table}
\centering
\small
\caption{  \small The average power consumption per second and overall operational duration of an AGR in different modes.}
\begin{tabular}{ccc}
\toprule
\textbf{Mode} & \textbf{Average Power} & \textbf{Time (mins)} \\
\midrule
Fly & 987.61 W  & 14  \\
\midrule
Hover & 532.07 W  & 26  \\
\midrule
Ground & 197.52 W  & 55  \\
\bottomrule
\end{tabular}
\label{table:1}
\end{table}

We evaluated the AGRNav's performance in 3 complex real-world environments where the robot's vision was obstructed by walls and bushes. In contrast to mapping-based methods that could result in potential collisions or suboptimal trajectories, our AGRNav reliably predicted hidden obstacles (Fig. 6a), enabling safer navigation and searching energy-efficient paths (Fig. 6b) through SCONet's completion capabilities. Additionally, it identified optimal landing areas by foreseeing unseen obstacles (Fig. 6c), with semantic information aiding in velocity optimization for shorter travel times. 

\begin{table}[htbp]
\caption{\small Comparison of published methods on the official SemanticKITTI benchmark.}
\begin{center}
\small
\setstretch{1.32}
\renewcommand{\arraystretch}{1.0}
\begin{tabularx}{\columnwidth}{@{}>{\centering\arraybackslash}p{2.8cm}*{5}{>{\centering\arraybackslash}X}@{}}  % Set the table to span the width of a single column
\toprule

\textbf{Method} & \textbf{\textit{IoU}} & \textbf{\textit{Prec.}} & \textbf{\textit{Recall}} & \textbf{\textit{FPS}} &  \textbf{\textit{mIoU}} \\
\midrule
SSCNet \textcolor{green}{\cite{song2017semantic}} & 53.20 & 59.13 & \textbf{84.15} & 12.00  & 14.55 \\
SG-NN \textcolor{green}{\cite{dai2020sg}} & 31.26 & 31.60 & 54.50 & 12.00  & 9.90 \\
J3S3Net \textcolor{green}{\cite{yan2021sparse}} & 51.10 & 40.23 & 61.09 & 1.73  & 23.80 \\
LMSCNet \textcolor{green}{\cite{roldao2020lmscnet}} & 54.89 & 82.21 & 62.29 & 18.50  & 14.13 \\
S3CNet \textcolor{green}{\cite{cheng2021s3cnet}} & 45.60 & 48.79 & 77.13 & 1.82 & \textbf{29.50} \\
TDS  \textcolor{green}{\cite{garbade2019two}}  & 50.60 & 72.43 & 78.61 & 1.70  & 17.70 \\
\textbf{SCONet (our)} & \textbf{56.12} & \textbf{85.02} & 63.47 & \textbf{20.00} & 17.61 \\
\bottomrule
\end{tabularx}
\label{tab_custom}
\end{center}
\end{table}

\subsection{Semantic Scene Completion Network (SCONet)}
\noindent\textbf{Pre-trained Model}. We evaluate SCONet's performance using the SemanticKITTI benchmark \cite{behley2019semantickitti}, which offers 3D voxel grids from HDL-64 LiDAR scans in urban settings, labelled semantically. The input and ground truth grids are sparse, with dimensions of $256 \times  256 \times  32$ and a $0.2 m$ voxel size. Our analysis concentrated on completion metrics (IoU, precision, recall) and the semantic metrics mIoU, utilizing the benchmark's original splits and enhancing generalization through $x-y$ flipping augmentation. Employing the Adam optimizer at a learning rate of 0.001, scaled by 0.98 per epoch and conducted on a machine with 4 NVIDIA RTX3090 GPUs, training achieved convergence within 24 hours.

\begin{figure}[t]
  \centering
     \includegraphics[width= \linewidth]{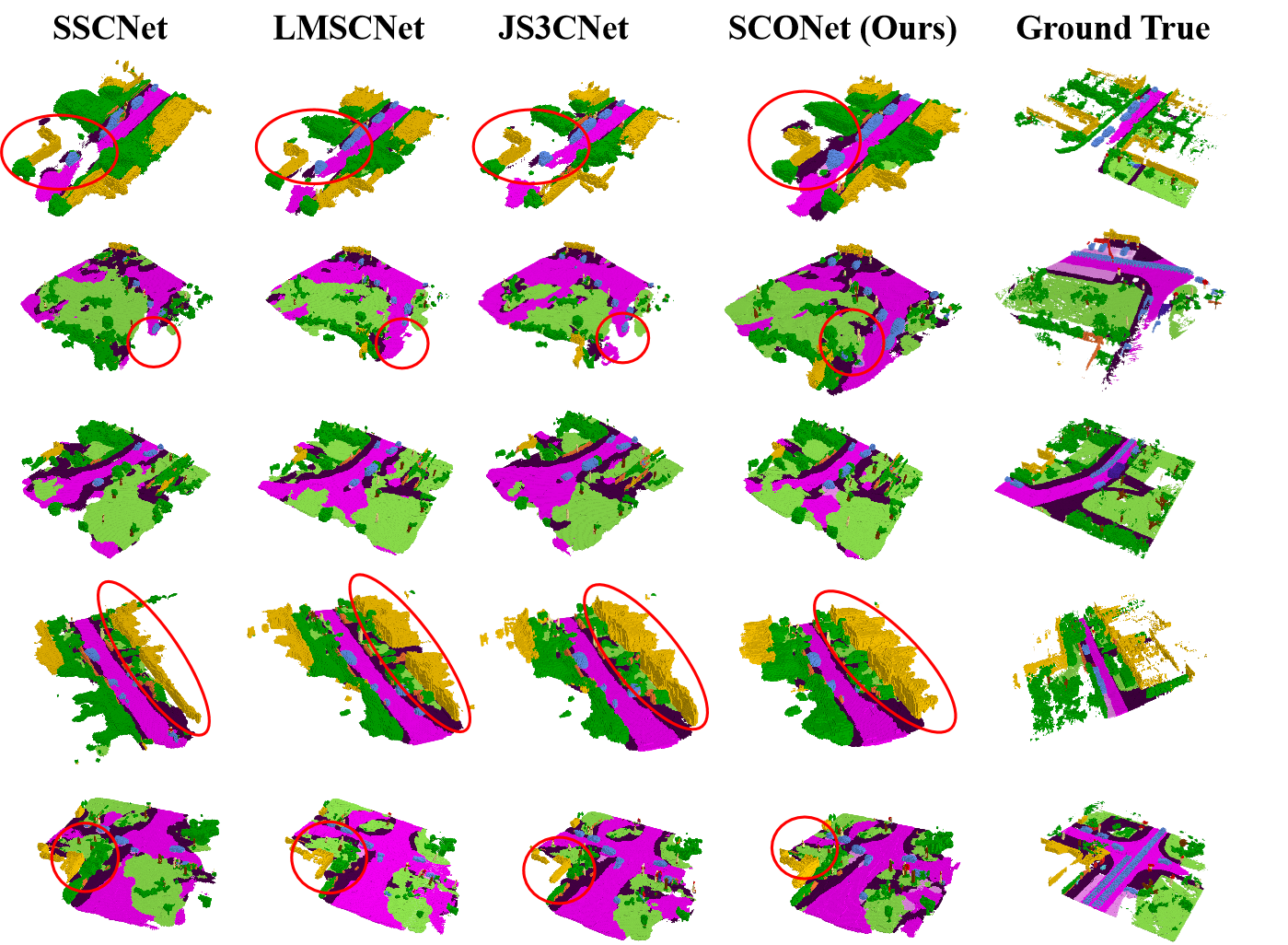}
   \caption{\small Qualitative results of SCONet on the validation set of SemanticKITTI. }
   \label{fig:onecol}
\end{figure}

\noindent\textbf{Quantitative Results.} Table 3 reveals that our SCONet outperforms its rivals, registering the highest IoU completion metric score of 56.12. This result stems from testing on a hidden dataset via the official server. Despite a slightly lower mIoU compared to S3CNet and J3S3Net, SCONet's inference speed is significantly enhanced, being about 20 times faster. This is primarily due to the adoption of depthwise separable convolutions instead of the resource-heavy 3D convolutions in its encoder, enabling real-time efficiency with 20 FPS on an RTX 3090 GPU.

\noindent\textbf{Qualitative Results.} Fig. 7 illustrates that SCONet outshines the baseline model, notably enhancing the completion of structural objects like vehicles and trees (row 5). It adeptly handles completion in areas obscured by trees or walls (rows 1 and 4), a key feature for enabling efficient and safe path planning in later stages.

\begin{table}
\centering

\caption{\small Ablation study of our model design choices on the SemanticKITTI \cite{behley2019semantickitti} validation set.}
\begin{tabular}{lll}
\toprule
Method & IoU $\uparrow$ & mIoU $\uparrow$ \\
\midrule
SCONet (ours) & 55.50 & 16.10 \\
w/o Depth-Separable Convolution & 54.15 & 15.76 \\
w/o Criss-Cross Attention & 53.20 & 15.11 \\
w/o MobileViT-v2 Attention & 53.86 & 15.37 \\
\bottomrule
\end{tabular}

\label{tab:example}
\end{table}

\noindent\textbf{Ablation experiment results.} Ablation studies on the SemanticKITTI validation set (Table 4) highlight the significance of two key components in our network: self-attention mechanisms and depth-separable convolutions. The CCA mechanism substantially impacts completion and semantic prediction by effectively aggregating context across rows and columns. \textit{Without CCA} causes a 4.14\% and 6.15\% drop for completion and semantic completion, respectively. Meanwhile, MobileViT-v2 Attention captures local scene features, such as occluded areas, with low computational overhead.  \textit{Without MobileViT-v2 Attention} leads to a 2.95\% decline in IoU. Furthermore, depth-separable convolutions significantly reduce the number of parameters.

\section{Conclusions}

In this paper, we introduce AGRNav, an efficient and energy-saving autonomous navigation framework for air-ground robots, featuring the key component SCONet, which outperforms state-of-the-art models in prediction accuracy and inference time. Additionally, a hierarchical path planner, improved by a query-based low-latency update method, considers obstacles in occluded areas to generate paths. This approach not only minimizes collision risk but also reduces energy consumption by 50\% compared to the baseline by cutting down high-energy aerial paths. The system's robustness has been extensively validated through experiments in both simulated and real-world environments.

\section*{ACKNOWLEDGMENT}
The work is supported in part by National Key R\&D Program of China (2022ZD0160200), HK RIF (R7030-22), HK ITF (GHP/169/20SZ), the Huawei Flagship Research Grants in 2021 and 2023, and HK RGC GRF (Ref: HKU 17208223), the HKU-SCF FinTech AcademyR\&D Funding Schemes in 2021 and 2022, and the Shanghai Artificial Intelligence Laboratory (Heming Cui is a courtesy researcher in this lab).
% %%%%%%%%%%%%%%%%%%%%%%%%%%%%%%%%%%%%%%%%%%%%%%%%%%%%%%%%%%%%%%%%%%%%%%%%%%%%%%%%

\bibliographystyle{IEEEtran}
\bibliography{references}

\end{document}